\documentclass[letterpaper, 10 pt, conference]{IEEEconf}  

\IEEEoverridecommandlockouts                              

\usepackage[pdftex]{graphicx} 
\usepackage{amsmath} 
\usepackage{amssymb}  
\usepackage{multirow}

\usepackage[utf8]{inputenc}

\usepackage{amsfonts}
\usepackage{float}
\usepackage{pifont}

\usepackage{algorithm}
\usepackage{algorithmic}

\usepackage{xcolor}

\usepackage{eurosym}
\usepackage{booktabs}
\usepackage{amstext} 
\DeclareRobustCommand{\officialeuro}{%
  \ifmmode\expandafter\text\fi
  {\fontencoding{U}\fontfamily{eurosym}\selectfont e}}

\title{\LARGE \bf
dPMP-Deep Probabilistic Motion Planning:\\ A use case in Strawberry Picking Robot
}

\author{Alessandra Tafuro$^{1}$, Bappaditya Debnath$^{2}$, Andrea M. Zanchettin$^{1}$,  and Amir Ghalamzan E.$^{3,\dagger}$
\thanks{ $^{\dagger}$ Corresponding author. Codes and dataset: {\tt github.com/imanlab/dPMP}}
\thanks{$^{1}$ Politecnico di Milano, Italy. $^{2}$ Edge Hill University, UK. $^{3}$ Lincoln Institute for Agri-food Technology, University of Lincoln, UK {\tt\small aghalamzanesfahani@lincoln.ac.uk}}
\thanks{*This work was partially supported by Centre for Doctoral Training, United Kingdom (CDT) in Agri-Food Robotics (AgriFoRwArdS) Grant reference: EP/S023917/1; Lincoln Agri-Robotics (LAR) funded by Research England; and by ARTEMIS project funded by Cancer Research UK C24524/A300038.}
}%

\begin{document}

\maketitle
\thispagestyle{empty}
\pagestyle{empty}

\begin{abstract}
This paper presents a novel probabilistic approach to deep robot learning from demonstrations (LfD). Deep movement primitives (DMPs) are deterministic LfD model that maps visual information directly into a robot trajectory. This paper extends DMPs and presents a deep probabilistic model that maps the visual information into a distribution of effective robot trajectories. {The architecture that leads to the highest level of trajectory accuracy is presented and compared with the existing methods.} Moreover, this paper introduces a novel training method for learning \emph{domain-specific latent features}. We show the superiority of the proposed probabilistic approach and novel latent space learning in the real-robot task of strawberry harvesting in the lab. The experimental results demonstrate that latent space learning can significantly improve model prediction performances. 
{The proposed approach allows to sample trajectories from distribution and optimises the robot trajectory to meet a secondary objective, e.g. collision avoidance. }
%
\end{abstract}

\vspace{-0.2cm}
\section{Introduction}
\label{sec:intro}
Learning from demonstrations (\textbf{LfD}) was proposed to reduce the programming cost and time of robots. 
However, LfD methods are still relying mostly on hand-designed or high-level features, e.g. via/start/terminal points, to be provided by human or Computer Vision ({CV}) libraries~\cite{tafuro2022}.  
 
\begin{figure}[t]
\vspace{.2cm}
  \centering
  \includegraphics[width=\linewidth]{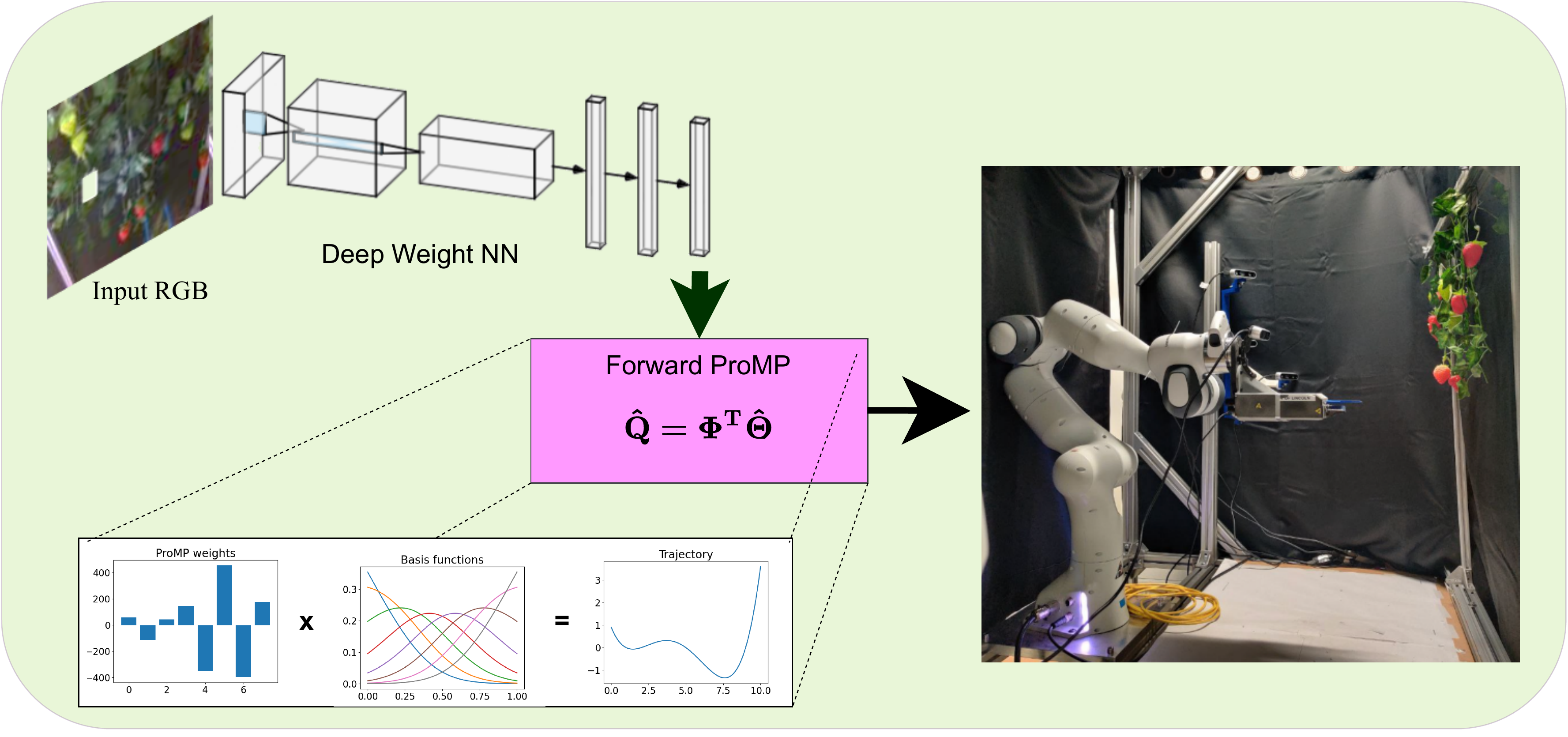}
  \caption{Schematic of Deep-MPs illustrates how the visual sensory information is translated into robot movements. This model is deterministic. Experimental setup is shown at right: a Franka arm sitting on a bench in front of plastic strawberries with camera-in-hand.}
 \vspace{-0.6cm}
\label{fig:panda_arm}
\end{figure}

Movement Primitives (\textbf{MP}s) are compact representations used for generating robot trajectory~\cite{ude2010task,paraschos2013probabilistic,mghames2020interactive}. Dynamic Movement Primitives (DMPs)~\cite{ude2010task} have been proved to be stable and robust against perturbations for discrete and rhythmic movement skills~\cite{matsubara2011learning}. 
Probabilistic Movement Primitives (\textbf{ProMPs})~\cite{paraschos2013probabilistic} is more suitable for planning robot movements that encode a distribution of demonstrated trajectories. ProMPs express the distribution of demonstrated trajectories by corresponding weights mean and variance. To capture the correlation between the visual information and the corresponding manipulation trajectories DMPs and ProMPs can be exploited.
Recent works~\cite{pervez2017learning,gams2018deep} proposed NN-based models trained to produce the  DMP parameters. 
Deep-MPs~\cite{sanni2022}, maps the raw image of the robot workspace into single robot trajectories (Deep-MPs). However,  Nonetheless, they are not suitable to generate a probabilistic distribution of trajectories may be inherent in human demonstrations. 
We extend Deep-MPs (see Fig.~\ref{fig:panda_arm}) to capture the variation in the demonstrations via the proposed probabilistic LfD approach. 
Ridge et al.~\cite{ridge2020training} proposed an NN-based model that was trained to produce the  DMP parameters \cite{schaal2006dynamic}.
In contrast to d-DMP \cite{pervez2017learning}\cite{ridge2020training}  and Deep-MPs, our approach maps the visual sensory information into a distribution of robot trajectories. 
Despite Deep-MPs which only generate one trajectory for a given image of the robot workspace, our proposed approach generates distributions of trajectories by the corresponding mean and variance. The approach is called Deep Probabilistic Motion Planning (dPMP).

In the basic dPMP, we have two independent training phases (\textbf{tp1}) unsupervised learning of latent space (e.g., training an auto-encoder (\textbf{AE})), and (\textbf{tp2}) supervised learning of the MPs weights. This is a shortcoming of dPMP and limits the performance of the model as the latent space is not \emph{domain}-specific (\emph{task}-specific). Hence, we present a 3 stages training that enforces the latent space to be domain-specific. After (tp1) and (tp2), we train the Encoder to further minimise the model loss, i.e. trajectory error making the latent features domain-specific (\textbf{tp3}). 

Encoder-Decoder is a well-known architecture in NNs, it first compresses the information, e.g., image, in spatial resolution and then gradually decompress it~\cite{badrinarayanan2017segnet}. In AE architecture, the output of the decoder is the same as the input of the Encoder. This compresses/embeds the feature representation into a lower-dimensional space (latent feature). This can be used for different tasks such as dimensionality reduction~\cite{wang2014generalized} and information retrieval~\cite{baldi2012autoEncoders}. AEs yield optimal reconstruction error but lack structure and interpretability of the latent space, i.e. it is not capable of generating new content. By regularising AEs during training and expressing the probabilistic latent space we get Variational AEs (\textbf{VAE})~\cite{an2015variational} to avoid over-fitting. Conditional VAE (\textbf{cVAE}) is introduced to learn the class-wise distribution of the data by inputting the condition (e.g., class label) into the latent space~\cite{kingma2013auto}. 
In dPMP, we compared AE, VAE, and cVAE to learn the low dimensional feature vector of the image describing the robot workspace. This latent vector is then mapped into the weights distribution of movement primitives which expresses the distribution of the demonstrated trajectories.  
AE, VAE, and cVAE are developed for CV-specific tasks. 
Hence, the latent space features generated by these methods are not robotic task-specific. 
For instance, the latent spaces learned for strawberry picking and strawberry plant pruning are identical. 
We are interested in robotic-task-specific latent space such that the latent space is tailored to contain more useful information for a specific task.

Our novel probabilistic approach to Deep LfD outperforms the state-of-the-art method in a series of real robot tasks of mock strawberry picking. 
The contributions of this paper include (1) a novel probabilistic model for dPMP using cVAE {that outperforms architectures with AE and VAE} (2) the implementation of a new training stage to learn domain/task-specific feature space. The proposed domain-specific training includes (1) unsupervised learning of initial latent space, (2) supervised learning for mapping latent feature vectors into robot trajectories distributions, (3) Encoder weights {tuning} using task-specific loss. We show the superiority and effectiveness of dPMP via a series of real robot tests for the simplified task of strawberry reaching with a Panda arm in the lab. The experimental results indicate that domain-specific training improves the model performances compared to others. 
\vspace{-0.25cm}

\section{Problem Formulation}

We consider a set of $N_{\mathrm{tr}}$ demonstrations $\mathcal{T} := \{\{\mathbf{Q}^1, \mathrm{I}^1\}, \ldots, \{\mathbf{Q}^{N_{\mathrm{tr}}}, \mathrm{I}^{N_{\mathrm{tr}}}\}\}$ where $\mathbf{Q}^n$, $n=1,\ldots,N_{\mathrm{tr}}$, are the joint/task space trajectories and $\mathrm{I}^{n}$ is raw RGB image taken from the corresponding robot's workspace (\textit{{\tt https://github.com/imanlab/dPMP}}). For a single joint, we define a trajectory as the ordered set $\mathbf{q} := \left\{q_t\right\}_{t=1,\ldots,T}$, where $q_t\in\mathbb{R}$ is the joint position at sample $t$, and $\mathbf{Q} := \{\mathbf{q}_1, . . ., \mathbf{q}_{N_{\mathrm{joint}}}\}$ where $N_{\mathrm{joint}}$ is the number of the joints (or Degree-of-freedoms (DOFs)) of a manipulator (or task).
Although we present the proposed approach in joint space, we can use the same step for task space.


\subsection{Learning from demonstration problem formulation}
We first model a robot trajectory with an observation uncertainty added to the following deterministic Movement Primitives (MPs) model:
\begin{equation}\label{eq:joint}
q = \sum_{i=1}^{N_{\mathrm{bas}}}\theta_i\psi_i(z(t)) + \epsilon_q
\end{equation} 
\noindent $\psi_i$ are basis functions (usually Gaussian~\cite{bishop2006pattern}) evaluated at $z(t)$. $z$ is a phase function that allows time modulation. If no modulation is required, then $z(t) = t/f$, where $f$ is the sampling frequency. $\theta_i\in\mathbb{R}$ are weights, and $\epsilon_q$ adds zero-mean Gaussian observation noise with variance $\Sigma_q$. 
We use normalised Gaussian basis functions for stroke-like movements 
$$ \psi_i(t):=\frac{b_i(z(t))}{\sum_{j=1}^{N_{\mathrm{bas}}}b_j(z(t))}$$ where $    b_i(z(t)):=\mathrm{exp}\left(-\frac{(z(t)-c_i)^2}{2h}\right)$. We can also write Eq. \eqref{eq:joint} in a matrix form $\mathrm{q}_t = \mathbf{\Psi}_t^{T}\mathbf{\Theta} + \epsilon_q$ where $\mathbf{\Psi}_t:=(\psi_1(z(t),\ldots,\psi_{N_{\mathrm{bas}}}(z(t))\in\mathbb{R}^{{N_{\mathrm{bas}}}\times 1}$, $\mathbf{\Theta}:=(\theta_1,\ldots,\theta_{N_{\mathrm{bas}}})\in\mathbb{R}^{{N_{\mathrm{bas}}}\times 1}$, and we also define $\mathbf{\Omega}:=(\mathbf{\Theta}_1,\ldots, \mathbf{\Theta}_{N_{\mathrm{joint}}})\in\mathbb{R}^{{N_{\mathrm{bas}}} N_{\mathrm{joint}} \times 1}$ and $\mathbf{\Phi}:=\left[\mathbf{\Psi}_1,\ldots,\mathbf{\Psi}_T\right]^T\in\mathbb{R}^{T\times N_{\mathrm{bas}}}$.

\emph{Deep Movement Primitives} (\textbf{deep-MPs}) introduced in \cite{sanni2022} uses an AI model learned from a task demonstrations (Fig.~\ref{fig:panda_arm}) to learn a  deterministic mapping from (visual) sensory information into the corresponding robot trajectory to complete a task. The MPs weight vector is then computed by a NNs, as per Eq.~\ref{eq:cov_mean_mp}: 
\begin{equation}    
    \boldsymbol{\Theta_j} = \mathrm{h}_j(\boldsymbol{W_j},\mathrm{I}^{n},\sigma_j)
        \label{eq:cov_mean_mp}
\end{equation}

We can express distribution of demonstrated trajectories similar to \emph{Probabilistic Movement Primitives} (\textbf{ProMPs})~\cite{paraschos2013probabilistic}.
From Eq. \eqref{eq:joint}, it follows that the probability of observing $q_t$ is given by $
    p(q_t|\mathbf{\Theta})=\mathcal{N}\left(q_t\,\big|\,\mathbf{\Psi}^{T}_t\mathbf{\Theta}, \mathbf{\Sigma}_q\right)$. 
Since $\Sigma_q$ is the same for every time step, the values $q_t$ are taken from independent and identical distributions, i.i.d. Hence, the probability of observing a trajectory $\mathbf{q}$ is given by:
\begin{equation}
    p(\mathbf{q}|\mathbf{\Theta}):=\prod_{t=1}^Tp(q_t|\mathbf{\Theta})
    \label{eq:joint_}
\end{equation}

We can also assume weight parameters are taken from a distribution. We estimate the distribution of $q_t$ which does not depend on $\mathbf{\Theta}$, but on $\rho:=(\mu_{\mathbf{\Theta}},\mathbf{\Sigma}_{\mathbf{\Theta}})$. This is done by marginalising $\mathbf{\Theta}$ out in the distribution as follows:
\begin{eqnarray}\label{eq:marginal}
p(q_t|\rho)&=&\int\mathcal{N}\left(q_t\,\big|\,\mathbf{\Psi}^{T}_t\mathbf{\Theta},\,\Sigma_q\right)\mathcal{N}\left(\mathbf{\Theta}\,\big|\,\mu_{\mathbf{\Theta}},\,\Sigma_{\mathbf{\Theta}}\right)\mathrm{d}\mathbf{\Theta}
\nonumber\\
&=&\mathcal{N}\left(q_t\,\big|\,\mathbf{\Psi}^{T}_t\mathbf{\Theta},\,\Sigma_q+\mathbf{\Psi}^{T}_t\mathbf{\Sigma}_{\mathbf{\Theta}}\mathbf{\Psi}_t\right)
\end{eqnarray}
where $\mathbf{\Theta}\sim p(\mathbf{\Theta}|\rho)=\mathcal{N}(\mathbf{\Theta}|\mu_{\mathbf{\Theta}},\mathbf{\Sigma}_{\mathbf{\Theta}})$. 

\begin{algorithm}[b!]
\caption{Domain-specific latent vector training dPMP.}
\label{algo1}
\begin{algorithmic}[1]
\item[{\textbf{Note}: This pseudo code is for joint trajectory prediction.}]\item[{Generalising it to the task space is straightforward.}]
\item[\textbf{Input}: MLPs architecture $h_j$, Encoder architecture] \item[$\mathit{Encoder}$, ProMPs basis functions $\mathbf{\Phi}$, image $\mathrm{I^n}$, training]
\item[set trajectories $\mathbf{q}_j,$ activation functions $\mathbf{\sigma}_j$ and $\mathbf{\sigma}_{enc}$.]
\item[{\textbf{Outputs}: Encoder and MLP weights $\mathbf{W}_{enc}$, $\mathbf{W}_{j}$, mean}]\item[{trajectory $\mathbf{\hat{q}}_{mean,j}$ and covariance matrix $\mathbf{\hat{\Sigma}}_{{q}_{,j}}$.}]
\item[     -------------------------------------------------------------------------]
\STATE $\mathit{Probabilistic}$ $\mathit{Dataset}$:\par\hspace{1cm} $\mathcal{T}= \{\mathit{\mathrm{q_j}, \mathrm{I^n}}\}_{n=\{1, . . .,N_{\mathrm{tr}}\},j=\{1, . . .,N_{joints}\}}$
\STATE $\mathit{Trajectories}$ $\mathit{Mean}$ $\mathit{and}$ $\mathit{Covariance}$ $\mathit{Extraction:}$
\par\hspace{1cm}$\{\mathbf{{q}}_{mean,j}$ , $\mathbf{{\Sigma}}_{{q}_{,j}}\}_{j=1...N_{joints}}$
\STATE $\mathit{Encoder}$-$\mathit{Decoder}$ $\mathit{training}$ $\mathit{for}$ $\mathit{image}$ $\mathit{reconstruction}$
\STATE $\mathit{Initialise}$ $\mathit{Deep Model:}$
\par\hspace{1cm}$\mathbf{E^n} = \: Encoder(\mathbf{W}_{enc},\mathbf{\mathrm{I^n}},\mathbf{\sigma}_{enc})$
\par\hspace{1cm}$\mathbf{\hat{\Theta}}_{mean,j},\mathbf{\hat{\Sigma}}_{\hat\Theta_j}\: =  \:h_j (\mathbf{W}_j, \mathbf{\mathbf{E^n}}, \mathbf{\sigma}_j)$
\STATE $\mathit{Initialise}$ $\mathit{ProMPs}$:  \par\hspace{1cm}$\mathbf{\hat{q}}_{mean,j} =\mathbf{\Phi}^T \mathbf{\hat{\Theta}}_{mean,j}$
\par\hspace{1cm}$\mathbf{\hat{\Sigma}}_{{q}_{,j}}= \mathbf{\Phi}^T \mathbf{\hat{\Sigma}}_{\hat\Theta,j}\mathbf{\Phi}$

\STATE $\mathit{RMSE}$ $\mathit{Loss:}$ Eq.~\ref{eq:RMSE_latent} 
\WHILE {$(\mathit{e} > \epsilon$)}
    \FORALL{$\{\mathit{\mathbf{q}_{mean,j},\mathbf{{\Sigma}}_{{q}_{,j}}, \mathrm{I^n}}\}\in  \mathcal{T}$}
        \STATE $\textsf{\textsc{Forward Propagation:}}$
        \par\hspace{0.5cm}$\mathbf{E^n} = Encoder(\mathbf{W}_{enc,j}, \mathbf{\mathrm{I^n}}, \mathbf{\sigma}_{enc,j})$
        \par\hspace{0.5cm}$\mathbf{\hat{\Theta}}_j,\mathbf{\hat{\Sigma}}_{\Theta_j} =  h_j (\mathbf{W}_j, \mathbf{\mathbf{E^n}}, \mathbf{\sigma}_j)$
        \STATE $\textsf{\textsc{Forward ProMPs:}}$  \par\hspace{0.5cm}$\mathbf{\hat{q}}_{mean,j} = \mathbf{\Phi}^T \mathbf{\hat{\Theta}}_{mean,j}$, \par\hspace{0.5cm}$\mathbf{\hat{\Sigma}}_{{q}_{,j}} = \mathbf{\Phi}^T \mathbf{\hat{\Sigma}}_{\hat\Theta,j}\mathbf{\Phi}$  
        \STATE $\textsf{\textsc{RMSE Loss:}}$\par\hspace{0.01cm}$ \: L_k =\begin{matrix} \sum_{j=1}^{N_{joints}} \| \mathbf{q}_{mean,j}(t_{end}) - \mathbf{\hat{q}}_{mean,j}(t_{end}) \|\end{matrix}$\par\hspace{1.8cm}$\alpha * \| \mathbf{\Sigma}_{{q}_{,j}}(t_{end}) - \mathbf{{\hat\Sigma}}_{{q}_{,j}}(t_{end}) \|$
    \ENDFOR
    \STATE $\textsf{\textsc{Back Propagation:}}$ $\mathit{Keep}$ $\mathbf{W}_j$  $Fixed$ $\&$ $\mathit{Train}$ $\mathbf{W}_{enc}$  $\mathit{with}$ $\mathit{Loss}$ $\mathit{in}$ $\mathit{Eq}$.~\ref{eq:RMSE_latent}
    \par\hspace{1cm}$\mathbf{W}_{enc}^{k+1} \gets  \{\mathbf{W}_{enc}^k, \frac{\partial{\mathit{L}_k}}{\partial{\mathbf{W}_{enc}^k}}\}$
     
\ENDWHILE

\STATE $\mathrm{dPMP}$: 
\par\hspace{1cm}$\mathbf{\hat{q}}_{mean,j} = \mathbf{\Phi}^T \mathbf{\hat{\Theta}}_{mean,j}$, \par\hspace{1cm}$\mathbf{\hat{\Sigma}}_{{q}_{,j}} = \mathbf{\Phi}^T \mathbf{\hat{\Sigma}}_{\hat\Theta,j}\mathbf{\Phi}$
\STATE \textbf{end}
\end{algorithmic}
\end{algorithm}

\begin{figure*}[tbh!]
\begin{center}
  \includegraphics[width=1\linewidth, trim={0 0 0 0cm },clip]{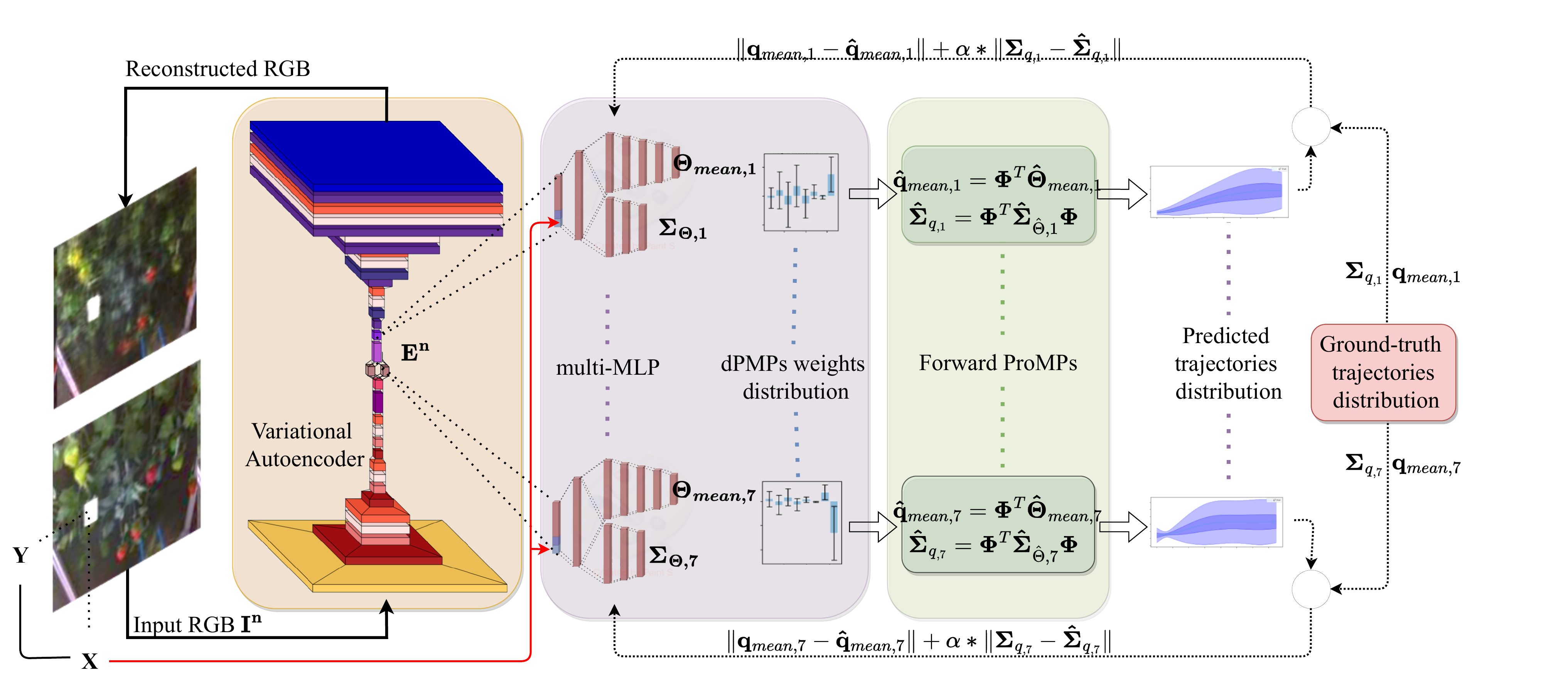}
  \includegraphics[width=1\linewidth, trim={0 0 181cm 0cm },clip]{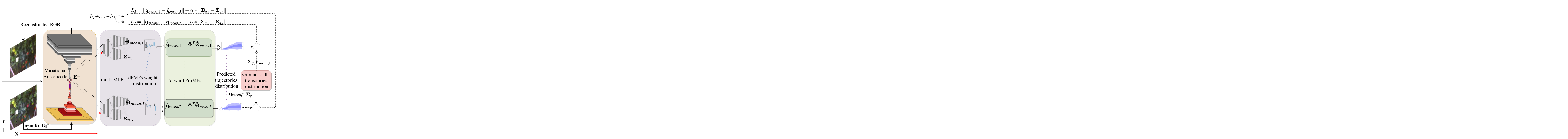}
\end{center}
 \caption{dPMP-cVAE: (top) shows the dPMP model architecture with conditional variational autoEncoder. The training has two phases (1) autoEncoder and (2) MLP training; (bottom) domain-specific latent space training: in addition to the two training phases above, (3) the weights of the Encoder are further trained by fixing the weights of MLPs and using a domain specific loss. {MLPs: a sequence of dense layers with 256 - 128 - 64 - 32 - 8 units predicting the ProMPs weights mean and/or covariance elements; Encoder: repetition of Convolutional layers with Leaky Relu activation function, and Max pooling and Batch normalization layers and a final Flattening layer. }}
 \vspace{-0.6cm}
\label{fig:model}
\end{figure*}

\subsection{Deep Probabilistic Motion Planning - dPMP}

\textbf{dPMP} is a probabilistic model that maps visual sensory information of a robot's workspace to the distribution of robot trajectories. 

For each joint, denoted by \textit{j}, the relative trajectory distribution can be expressed in weights space with a mean vector $\boldsymbol{\Theta_{mean,j}}$ and a covariance matrix $\boldsymbol{\Sigma_{\boldsymbol{\Theta}_,j}}$. The deep model can learn the relation between these two parameters an the input image.
\begin{equation}    
    \boldsymbol{\Theta_{mean,j}},\boldsymbol{\Sigma_{\boldsymbol{\Theta}_{,j}}} = \mathrm{f}_j(\boldsymbol{\hat{W}_j},\mathrm{I}^{n},\hat{\sigma}_j)
        \label{eq:cov_mean_ProMPs}
\end{equation}
We do not consider the correlation among the weights of different joints to reduce the dimensional of the problem. Considering these correlations can be a future work.  
Eq.~\eqref{eq:cov_mean_ProMPs} shows the network target vectors $\boldsymbol{\Theta_{mean,j}}$, $\boldsymbol{\Sigma_{\boldsymbol{\Theta}_{,j}}}$ are equivalent to a non-linear function, denoted by $\mathrm{f_j}$, of the input image $\mathrm{I^n}$,the weight parameter $\boldsymbol{\hat{W}_j}$ and the node activation $\mathrm{\hat{\sigma}_{j}}$. $\mathrm{f_j}$ is a non-linear deep model mapping the image $\mathrm{I^n}$ taken by robot’s camera at a home position (see Fig. \ref{fig:panda_arm}) to the ProMPs weights distributions of each joint. The weights distribution then generates the corresponding trajectories distribution {(\textit{Algorithm 1: Forward ProMPs})} using Eq.~\eqref{eq:traj_cov_mean}.
 \begin{equation}
\vspace{-0.3cm}
\begin{gathered}    
   \mathbf{\hat{q}}_{mean,j} = \mathbf{\Phi}^T \mathbf{\hat{\Theta}}_{mean,j}, \\ \mathbf{\hat{\Sigma}}_{{q}_{,j}} = \mathbf{\Phi}^T \mathbf{\hat{\Sigma}}_{\hat\Theta,j}\mathbf{\Phi}
        \label{eq:traj_cov_mean}
\end{gathered}
 \end{equation}

\vspace{0.3cm}
\subsubsection{Auto-Encoder (AE) and Variational AE (VAE),  for learning weights of dPMP}
We proposed and tested different baselines for dPMP model architectures to improve the accuracy in the prediction of the demonstrated behavior. 
The models have two parts {(\textit{Algorithm 1.9: Forward Propagation})}: (1) part-1 encodes the high-dimensional input RGB image in a low-dimensional latent space vector, preserving all the relevant information, using a set of convolutional layers, as per Eq.~\eqref{eq:encoding}; 
\begin{equation}    
\mathbf{E^n}= Encoder(\mathbf{W}_{enc}, \mathbf{\mathrm{I^n}}, \mathbf{\sigma}_{enc})
        \label{eq:encoding}
\end{equation}
(2) part-2 maps the latent embedded vector to the relative ProMPs weights distribution using a Multi Layer Perceptron (MLP) per each joint as per Eq.~\eqref{eq:MLP}. We solve the reduced dimensionality problem by using multiple MLPs (one for each joint). In future works, we will investigate using a single MLP that calculates the mean vector and variance across all the joints. 
\begin{equation}    
\mathbf{\hat{\Theta}}_j,\mathbf{\hat{\Sigma}}_{\Theta_j} =  h_j (\mathbf{W}_j, \mathbf{\mathbf{E^n}}, \mathbf{\sigma}_j)
        \label{eq:MLP}
\end{equation}
The twofold design of the model has an advantage over the direct mapping of the image to the trajectories distributions (as in~\cite{sanni2022}) as it yields higher accuracy as it is demonstrated in Sec.~\ref{sec:experiments}.
Our first baseline uses the encoding layers (Encoder) of an AE for part-1 of the dPMP model to reduce the input dimensionality while preserving the important information. We first train an AE using images of our dataset in which the decoder reconstructs the input RGB image from the latent representation to compute the reconstruction loss. 
In the second baseline, we used tha Encoder of a VAE so the latent representation $\mathbf{E^n}$ is stochastic as per Eq.~\eqref{eq:VAE}. {Both the AE and VAE are made of a sequence of convolutional layers with Leaky Relu activation function, Max pooling, Batch normalization layers and a final Flattening layer.} 
The Part-2 of the model {is a multi layer perceptron (MLP)}, so a sequence of dense layers that map the deterministic or stochastic latent representation of the input image to the ProMPs weights distributions as per Eq.~\eqref{eq:MLP} .
\begin{equation}
\begin{gathered} 
\mathbf{E^n}\sim\mathcal{N}(\mathbf{\mu}_{\boldsymbol{E^n}},{\Sigma}_{\boldsymbol{E^n}})\\
\mathbf{\mu}_{\boldsymbol{E^n}},{\Sigma}_{\boldsymbol{E^n}}= \mathit{Encoder}(\mathbf{W}_{enc}, \mathbf{\mathrm{I^n}}, \mathbf{\sigma}_{enc})
\label{eq:VAE}
\end{gathered}
 \end{equation}

\subsubsection{Conditioned dPMP model}
In the third variation of the model architecture, we add a conditional variable $\mathbf{c}$ concatenated with the latent vector computed by the Encoder from the input image. This helps the consequent MLP to tailor its behaviour according to some abstract information, e.g., pixel coordinate of a target berry, as per Eq.~\eqref{eq:CVAE}.
\begin{equation} 
\mathbf{\hat{\Theta}}_j,\mathbf{\hat{\Sigma}}_{\Theta_j} =  h_j (\mathbf{W}_j, \mathbf{\mathbf{E^n}}, \mathbf{\sigma}_j,\mathbf{c})
        \label{eq:CVAE}
\end{equation}

\subsubsection{Learning domain-specific latent space}
\begin{figure*}[t!]
  \includegraphics[width=.85\linewidth]{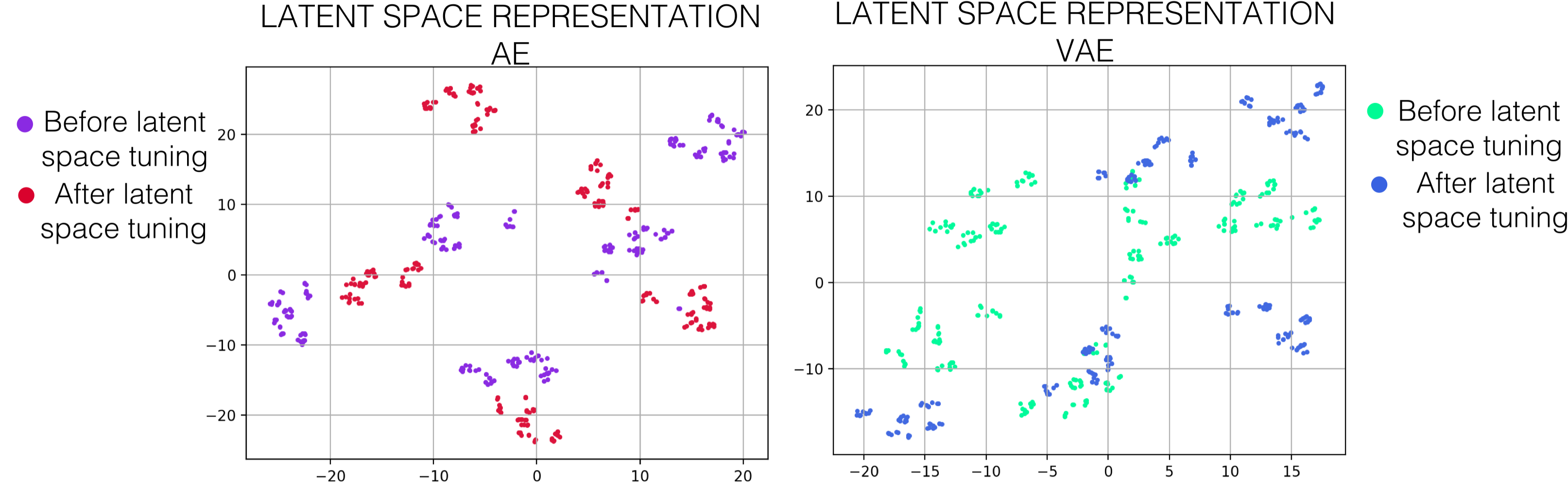}
  \centering
  \caption{Latent space visualisation/representation through T-SNE embedding before and after domain specific tuning.  After tuning, the clustering in the embedding space increases in both AE and VAE. A numerical evaluation is present in Table \ref{table:latent}.}
\label{fig:latent_Space}
\end{figure*}

Of note, we have two ordered training stages for the previous models: (tr-1) unsupervised training of part-1 (using image reconstruction loss) and (tr-2) supervised training of part-2 (using RMSE error as per Eq.~\eqref{eq:RMSE} where $\alpha$ is a tuning parameter that weights the loss components).
\begin{equation} 
\mathit{e} =  \| \mathbf{{q}}_{mean,j} - \mathbf{\hat{q}}_{mean,j} \| + \alpha  \| \mathbf{\Sigma}_{{q}_{,j}} - \mathbf{{\hat\Sigma}}_{{q}_{,j}} \|
        \label{eq:RMSE}
\end{equation}
Eq. \eqref{eq:RMSE} indicates the loss is computed after having reconstructed the mean and covariance matrix of the demonstrated trajectories computed by ProMPs weights distribution as per Eq. \eqref{eq:traj_cov_mean}. We first do tr-1 and train the Encoder; then, we do tr-2 to train the MLPs. Hence, tr-1 and tr-2 are completely decoupled. As such, the latent space is defined to maintain the information necessary for reconstructing the input RGB image, independently from the demonstrated trajectories distributions. 
We argue that, while this non-domain-specific training is absolutely useful for CV, it is not relevant for robotic tasks. Hence, we propose to train the weights of Encoder using the loss in Eq.~\ref{eq:RMSE_latent} while the MLPs weights are kept fixed (Fig.~\ref{fig:latent_Space}). 
In this way, there is also a direct mapping of the latent space to the information useful both for image reconstruction and trajectory prediction. {The dPMPs model obtained after the domain-specific latent space tuning is indicated with the \textit{l} prefix (l-dPMP).} The domain-specific latent space learning dPMP is explained in detail in \emph{Algorithm 1}.

\begin{flushleft}
\begin{table}[!t]
\vspace{.3cm}
\centering
\caption{Joint space predictions.\\ deep-ProMp-cVAE is the most accurate model.}
\label{table:joint}
\begin{tabular}{@{}cccccc@{}}
\multirow{2}{*}{Joint} & deep-ProMP-AE & \multicolumn{2}{c}{deep-ProMP-VAE} & \multicolumn{2}{c}{deep-ProMP-cVAE}\\
& RMSE & RMSE & drop  & RMSE & drop\\
\toprule
J1 &  5.4$\times 10^{-4}$  & 1.3$\times 10^{-4}$ & -74\% & 1.0$\times 10^{-4}$ & -28\%\\
J2 &  48.9$\times 10^{-4}$ & 14.4$\times 10^{-4}$ & -70\% & 9.0$\times 10^{-4}$ & -38\% \\
J3 &  93.0$\times 10^{-4}$ & 9.6$\times 10^{-4}$ & -89\% & 5.9$\times 10^{-4}$ & -38\% \\
J4 &  37.1$\times 10^{-4}$ & 19.3$\times 10^{-4}$ & -47\% & 5.9$\times 10^{-4}$ & -71\% \\
J5 &  24.2$\times 10^{-4}$ & 23.5$\times 10^{-4}$ & -2.7\% & 20.1$\times 10^{-4}$ & -13\% \\
J6 &  29.0$\times 10^{-4}$ & 15.4$\times 10^{-4}$& -48\% & 15.4$\times 10^{-4}$ & -0\% \\
J7 &  21.4$\times 10^{-4}$ & 5.4$\times 10^{-4}$ & -75\% & 5.4$\times 10^{-4}$ & -0\% \\
\bottomrule
\end{tabular}
\end{table}
\end{flushleft}

\begin{flushleft}
\begin{table}[tb!]
\centering
\caption{Cluster separability of latent space representation through T-SNE using Davies-Bouldin Index before and after latent space tuning with . A lower score indicates better clustering. Fig. \ref{fig:latent_Space} shows the T-SNE visualisations}
\label{table:latent}
\begin{tabular}{cc|c}
\multicolumn{1}{l}{} & \multicolumn{1}{l|}{} & \multicolumn{1}{l}{\textbf{Davies-Bouldin score}} \\
\multicolumn{1}{c|}{\multirow{2}{*}{AE}}  & base               & 0.616 \\
\multicolumn{1}{c|}{}                     & \textbf{latent space tuned} & $\mathbf{0.482}$ \\ \hline
\multicolumn{1}{c|}{\multirow{2}{*}{VAE}} & base               & 0.424 \\
\multicolumn{1}{c|}{}                     & \textbf{latent space tuned} & $\mathbf{0.352}$
\end{tabular}
\end{table}
\end{flushleft}

\begin{table*}[!tbh]
\centering
\vspace{.3cm}
\caption{Joint space vs Task Space prediction accuracy at the final point. Task space predictions are more accurate.}
\label{table:joint_vs_task}
\begin{tabular}{ccccc|cccc}
 &
  \multicolumn{4}{c|}{joint space prediction} &
  \multicolumn{4}{c}{task space prediction} \\ \cline{2-9} 
\multicolumn{1}{c|}{} &
  \multicolumn{2}{c|}{mean} &
  \multicolumn{2}{c|}{3std.} &
  \multicolumn{2}{c|}{mean} &
  \multicolumn{2}{c|}{3std.} \\
\multicolumn{1}{c|}{} &
  position [-] &
  \multicolumn{1}{c|}{orientation [-]} &
  position [-] &
  orientation [-] &
  position [-] &
  \multicolumn{1}{c|}{orientation [-]} &
  position [-] &
  \multicolumn{1}{c|}{orientation [-]} \\ \hline
\multicolumn{1}{c|}{deep-ProMP-Direct} &
  0.15 &
  \multicolumn{1}{c|}{0.1} &
  0.31 &
  0.41 &
  0.08 &
  \multicolumn{1}{c|}{0.06} &
  0.10 &
  \multicolumn{1}{c|}{0.10} \\
\multicolumn{1}{c|}{deep-ProMP-AE} &
  0.10 &
  \multicolumn{1}{c|}{0.06} &
  0.32 &
  0.41 &
  0.04 &
  \multicolumn{1}{c|}{0.04} &
  0.09 &
  \multicolumn{1}{c|}{0.07} \\
\multicolumn{1}{c|}{deep-ProMP-VAE} &
  0.06 &
  \multicolumn{1}{c|}{0.03} &
  0.19 &
  0.31 &
  0.03 &
  \multicolumn{1}{c|}{0.03} &
  0.07 &
  \multicolumn{1}{c|}{0.06} \\
\multicolumn{1}{c|}{deep-ProMP-cVAE} &
  0.04 &
  \multicolumn{1}{c|}{0.02} &
  0.18 &
  0.28 &
  0.02 &
  \multicolumn{1}{c|}{0.02} &
  0.06 &
  \multicolumn{1}{c|}{0.06}
\end{tabular}
\end{table*}
\begin{flushleft}
\begin{figure}[tb]
\begin{center}
  \includegraphics[width=1\linewidth]{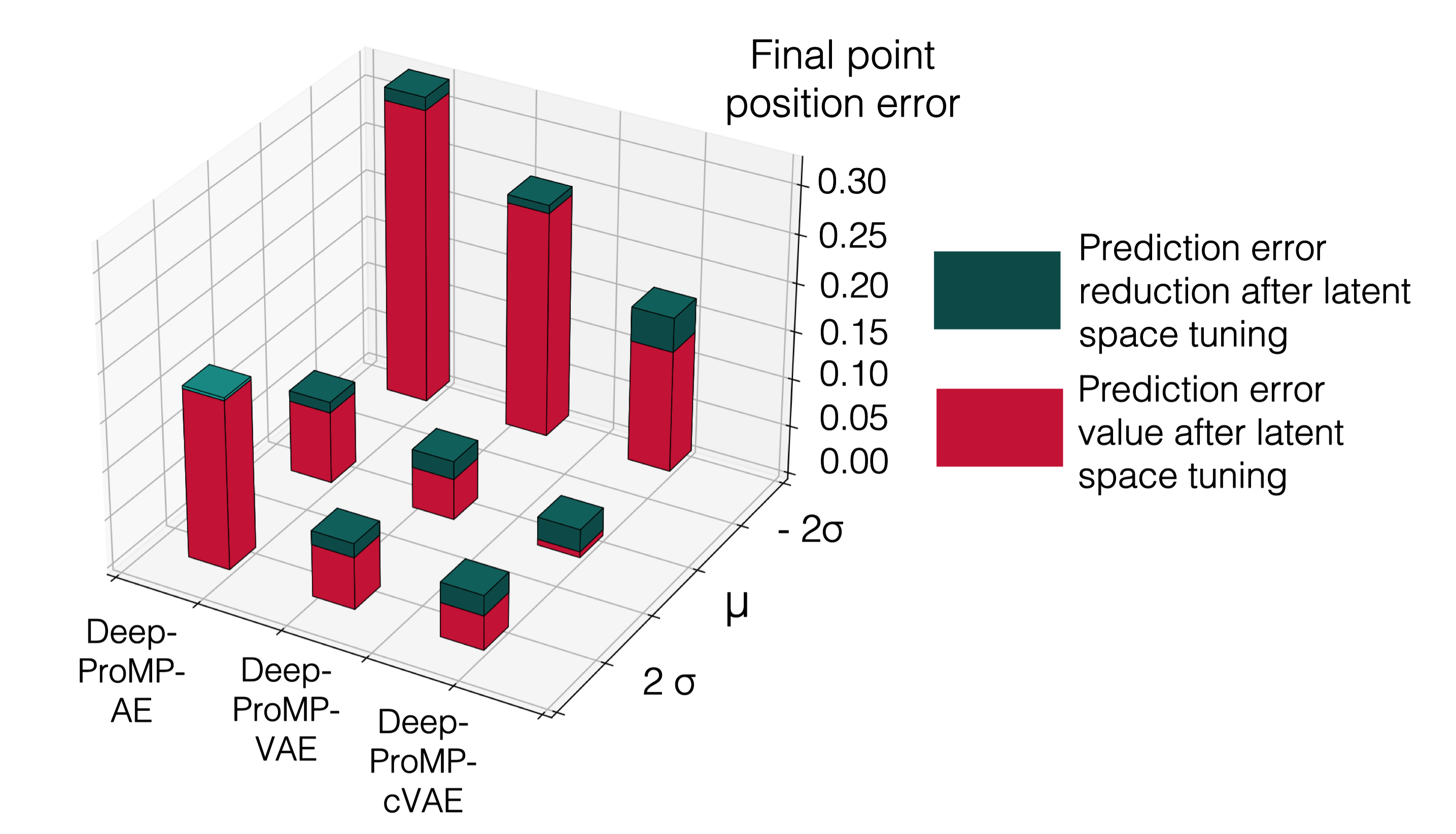}
\end{center}
  \caption{Experimental results from the implementation on the real robot. It is visible that performances increase going from Deep-ProMP-AE to deep-ProMP-VAE to Deep-ProMP-cVAE. A further improvement can be noticed after the domain-specific training of the encoders. Finally it is evident that the mean trajectory leads to higher precision in the final point with respect to the other sampled trajectories. }
\label{fig:results}
\end{figure}
\end{flushleft}
\vspace{-1.3cm}
\section{EXPERIMENTAL RESULTS AND DISCUSSION}
\label{sec:experiments}

\paragraph{Use case on selective harvesting of strawberries}
Developing robotic technologies for selective harvesting of high-value crops, such as strawberries, has been highly demanded because of different social, political, and economical factors \cite{duckett2018agricultural}, such as labour shortage and the recent COVID-19 pandemic. 
As such, private and public sectors have invested across the globe to develop robotic harvesting and commercialise the corresponding technologies in the last few years. 
Strawberry is among high-value crops with considerable selective harvesting cost in production (\textdollar 1 Billion is only the picking cost \cite{pickingcost21}). Hence, development of a commercial autonomous system for strawberry harvesting is of great interest.

\paragraph{Experimental setup and data set acquisition} To validate the proposed models {and select the best architecture}, we performed a series of real-robot experiments of strawberry approaching with a mock setup in the laboratory (Fig.~\ref{fig:panda_arm}). The experimental setup consists of a 7-DoF Panda robotic arm manufactured by Franka Emika with a custom gripper specifically designed for strawberry picking~\cite{vishnu2022}. An Intel RealSense D435i RGB-D camera is mounted on the top of the gripper. The images are captured with VGA resolution (640,480). We build a mock setup with plastic strawberries to test this method during the off-harvesting season. Nonetheless, we aim to test this approach in the field in future research stages. 
We collected a dataset of 250 samples where each sample includes an RGB image of the scene and robot trajectory starting from a home configuration as shown in Fig.~\ref{fig:panda_arm}. After taking an image, the robot is manually moved to reach the targeted berry in kinesthetic teaching mode. Although, we tested the proposed approach in a simple point-to-point movement, this encodes the expert knowledge in a complex corresponding task of strawberry picking where the robot needs to eventually interact with a dense cluster of strawberries. We mask the target berry in the input RGB image with a white bounding box using a trained segmentation model based on Detectron-2 \cite{wu2019detectron2}. {In the real application the target berry could be selected in different ways e.g. an algorithm schedules the picking order. }
The movement is repeated 10 times for a single targeted strawberry to capture the demonstration variations. We recorded the joint space and task space trajectories. We created 5 different strawberry plant configurations, each including 5 target berries. 

\paragraph{Implementation details} We trained and tested the different architectures for trajectory prediction both in the task and joint space. The first two models embed the input image with the Encoder of AE (dPMP-AE) or of VAE (dPMP-VAE) respectively. In both models, the latent space representation is a 256-d vector. In the third model, we condition a VAE by concatenating the pixel coordinate of the target berry bounding box center and VAE latent vector (dPMP-cVAE).
Moreover, we implemented domain-specific latent space learning of these three models (l-dPMP-AE, l-dPMP-VAE, l-dPMP-cVAE). We use the prediction error in the last time instant for the latent space tuning loss {(\textit{Algorithm 1.11: RMSE Loss})}, as per Eq.~\eqref{eq:RMSE_latent}. This allows tuning the Encoder weights according to specific task requirements. 

\begin{equation}
\begin{gathered}
    \mathit{e} =  \| \mathbf{{q}}_{mean,j}(t_{end}) - \mathbf{\hat{q}}_{mean,j}(t_{end}) \| +\\
    \alpha \| \mathbf{\Sigma}_{{q}_{,j}}(t_{end}) - \mathbf{{\hat\Sigma}}_{{q}_{,j}}(t_{end}) \|
    \label{eq:RMSE_latent}
\end{gathered}
\end{equation}

\paragraph{Discussion} we carried out three types of analysis and the corresponding results are respectively shown in Table \ref{table:joint_vs_task}, Table \ref{table:joint} and Fig. \ref{fig:results}. 
For the first test, we use the prediction performances of the 7 joint trajectories distributions with dPMP-AE, dPMP-VAE, and dPMP-cVAE. 
{We compared them to dPMP-Direct which is a model that maps directly the input image to the trajectories distributions with a series of Convolutional layers with Leaky Relu activation function, Max pooling, Batch normalization layers and a final Flattening layer followed by some Dense layers}. We use dPMP-Direct as a benchmark to demonstrate that the twofold design proposed performs better than the direct mapping. 
The performances have been evaluated on a test set never seen in training or validation stages. We used as evaluation metric the loss used for training, as shown in eq. \ref{eq:RMSE}. Table \ref{table:joint} shows that the prediction error is improving from dPMP-Direct to dPMP-AE, dPMP-VAE, and dPMP-cVAE. Hence, dPMP-cVAE is outperforming all other models. The same observation can be done in the task space case.
Another experimentation has been done on the same test set to compare the model performances between task and joint space predictions. The evaluation metric used is expressed in Eq.~\ref{eq:posorierror} considering the error at the final point (when the target berry has to be reached) between the predicted and ground truth end-effector position and orientation.

\begin{equation}
\begin{gathered} 
e_{position} =\sqrt{(x-\hat{x})^2+(y-\hat{y})^2+(z-\hat{z})^2}\\
e_{orientation} = min[\|q-\hat{q}\|,\|q+\hat{q}\|]
\label{eq:posorierror}
\end{gathered}
\end{equation}
In Eq. ~\ref{eq:posorierror}, $(x,y,z)$ and  $(\hat{x},\hat{y},\hat{z})$, and $q$ and  $\hat{q}$ represent the ground truth and predicted positions and orientation (quaternions) at the final time step of the end effector, respectively. {The error on the orientation has been derived from \cite{article} and expresses the distance between two quaternions.}
The trajectories that have been evaluated are the predicted mean trajectory and the one at a distance of $3\sigma$ from the mean. 
Table \ref{table:joint_vs_task} illustrates that task space predictions produce much more accurate performances. Moreover, the most accurate model is confirmed as being the  dPMP-cVAE.
We performed a third set of experiments by implementing the models on the real robot instead of using a previously recorded set of data. 
We built 5 new (different) strawberry plant configurations each including again 5 different target berries. 
This helps to demonstrate the generalisation ability of the model in predicting the reaching movement in unseen real-robot settings. 
The final position of the target strawberry is used as a reference to evaluate the prediction performances. To capture this information the robot was moved in kinesthetic teaching mode to the desired final pose necessary for picking a target berry and the final $(x,y,z)$ position was recorded. 
dPMP-AE, dPMP-VAE, dPMP-cVAE have been tested before and after the domain-specific latent space training. The predicted mean trajectory together with the trajectories at $2\sigma$ and at $-2\sigma$ from the mean have been evaluated. We use Eq.~\eqref{eq:posorierror} as the metric to evaluate the position error at the final reaching point. 
Fig. \ref{fig:results} shows that the model performances increase after the domain-specific training. Furthermore, the most accurate model is l-dPMP-cVAE. 
The probabilistic framework can be exploited to perform the task in different ways sampling from the predictions of dPMP. For example, this allows the robot to reach the target point with different orientations.
We also studied the clustering level of the latent space before and after the domain-specific training. 
Fig.~\ref{fig:latent_Space} visualises the 2-D representation of latent spaces of AE Encoder or VAE Encoder to embed the input images (in a 256-d latent vector) before and after the domain-specific training. 
T-distributed Stochastic Neighbour Embedding \cite{belkina2019automated} was used to reduce the dimensionality of the latent vectors from 256 to 2 to visualise the latent vectors. Every point in this representation in Fig.~\ref{fig:latent_Space} represents an image taken from the dataset of 250 RGB images used for model training and testing. 
Table \ref{table:latent} also shows the Davies-Bouldin score used to evaluate the clustering level of the latent space with a numeric index. 
The score indicates the average similarity measure of each cluster with its most similar cluster, where similarity is the ratio of within-cluster distances to between-cluster distances. Thus, the lower the scores, the higher the level of clustering. 
The clustering level increases from AE to the VAE. Moreover, the domain-specific latent space learning further increases the clustering level, teams
{which is a well-known sign of better representation of the distribution of the inputs}.

{Similar to ProMPs features, such as conditioning by via-points, final positions or velocities, time modulation, and blending of movement primitives, we will investigate these features in our future works. Moreover, we will study how to adopt Kullback-Leibler divergence as the loss to train part-2 of the model since it explicitly expresses the distance between 2 probability distributions. 
We will also consider the use of dPMP in teleoperation setup to reduce the cognitive workload on the human operator \cite{parsa2020haptic}.}
\section{CONCLUSIONS}
This paper presents a novel probabilistic framework for deep movement primitives which maps the visual sensory information of a robot workspace into the corresponding robot trajectories according to the demonstrations of the desired task. 
{We propose a model architecture that leads to accurate predictions, namely dPMP-cVAE, found trough its comparison with dPMP-AE, dPMP-VAE, and dPMP-Direct}. We demonstrated the effectiveness of the proposed dPMP in a series of real robot tests showing the robot can produce a distribution of trajectories for a a single input image enabling the robot to access infinite number of trajectories that can be sampled from the distribution. 
To further improve the performance of dPMP, we proposed a novel domain-specific latent space tuning method that allows learning the latent space representation in order to be relevant for the specific robotic task. The results suggest this model, namely l-dPMP-cVAE, outperforms the other proposed models {with acceptable accuracy for real robot applications.} 




\bibliographystyle{IEEEtran}

\bibliography{IEEEexample,egbib}

\end{document}